\documentclass[runningheads]{llncs}
\usepackage{graphicx}
\usepackage{amsmath,amssymb} 
\usepackage{tikz}

\usepackage{color}

\begin{document}

\newcommand{\circlemarker}{
\hspace*{-5pt}
\begin{tikzpicture}
\draw[black,fill=black] (0,0) circle (.75ex);
\draw[white,fill=white] (0,0) circle (.35ex);
\end{tikzpicture}
\hspace{-1pt}
}

\newcommand{\boxmarker}[1]{
\hspace*{-8pt}
\begin{tikzpicture}
\draw[#1,fill=#1] (0,0) rectangle (1.25ex,1.25ex);
\end{tikzpicture}
\hspace*{-5pt}
}

\definecolor{colorRLA}{rgb}{0.6,0.247,0.0}
\definecolor{colorRUA}{rgb}{0.0,0.459,0.863}
\definecolor{colorLLA}{rgb}{1.0,0.314,0.02}
\definecolor{colorLUA}{rgb}{1.0,0.643,0.02}
\definecolor{colorRUL}{rgb}{0.0,0.361,0.192}
\definecolor{colorLUL}{rgb}{1.0,1.0,0.0}
\definecolor{colorBlack}{rgb}{0.0,0.0,0.0}
\definecolor{colorRed}{rgb}{0.8500, 0.3250, 0.0980}
\definecolor{colorBlue}{rgb}{0.0, 0.4470, 0.7410}

\title{Multi-Kernel Diffusion CNNs for Graph-Based Learning on Point Clouds}
%
%
\author{Lasse Hansen\inst{1} \and Jasper Diesel\inst{2} \and Mattias P. Heinrich\inst{1}}
\authorrunning{L. Hansen et al.}
%
\institute{Institute of Medical Informatics, University of L\"ubeck, DE \and Dr\"ager, L\"ubeck, DE\\
\email{hansen@imi.uni-luebeck.de}
}

\maketitle              
\begin{abstract}
Graph convolutional networks are a new promising learning approach to deal with data on irregular domains. They are predestined to overcome certain limitations of conventional grid-based architectures and will enable efficient handling of point clouds or related graphical data representations, e.g. superpixel graphs. Learning feature extractors and classifiers on 3D point clouds is still an underdeveloped area and has potential restrictions to equal graph topologies. In this work, we derive a new architectural design that combines rotationally and topologically invariant graph diffusion operators and node-wise feature learning through $1\times1$ convolutions. By combining multiple isotropic diffusion operations based on the Laplace-Beltrami operator, we can learn an optimal linear combination of diffusion kernels for effective feature propagation across nodes on an irregular graph. We validated our approach for learning point descriptors as well as semantic classification on real 3D point clouds of human poses and demonstrate an improvement from 85\% to 95\% in Dice overlap with our multi-kernel approach.

\end{abstract}

\section{Introduction}

The vast majority of image acquisition and analysis has so far focused on reconstructing and processing dense images or volumetric data. This is mainly motivated by the simplicity of representing data points and their spatial relationships on regular grids and storing or visualising them using arrays. In particular convolutional operators for feature extraction and pooling have seen increased importance for denoising, segmentation, registration and detection due to the rise of deep learning techniques. Learning spatial filter coefficients through backpropagation is well understood and computationally efficient due to highly optimised matrix multiplication routines for both CPUs and GPUs.

However, many alternative imaging devices such as time-of-flight based 3D scanners or ultrasound that is based on reflectance measurements are not necessarily optimally represented on dense 3D grids. Instead these sparse measurements can be stored and processed more naturally and effectively using point clouds that are connected by edges forming an irregular graph. Moreover, 3D data from multiple sources can be easily combined if represented as point clouds.

The supervised feature learning and further analyses on these irregular domains is a research area that is still in its early stage, in particular in the context of deep learning. The main limitations of previous approaches are their dependency on an equal number of nodes in all graphs (e.g. derived from point clouds) and the same topology, i.e. ordering of nodes and edge connections. Furthermore, some operations on irregular graphs are inefficient for parallel hardware, which limits their usefulness in real world scenarios.

\subsection{Related Work}

Of all hierarchical feature learning models, convolutional neural networks have shown to be one of the most successful approaches for a wide variety of tasks~\cite{Ren2017}. Attempts to transfer the concepts from the two dimensional image domain directly to a sparsely sampled 3D space include e.g. volumetric CNNs \cite{Maturana2015} and multi-view CNNs \cite{Su2015}. However, due to the sparseness of the observed space both techniques lack computational efficiency.

Another class of works addresses this problem more generally by studying the intrinsic structure of data on non-Euclidean and irregular domains. Noteworthy are in particular spectral descriptors that are based on the eigenfunctions and eigenvalues of the Laplace-Beltrami operator. The proposed methods include heat kernel signatures (HKS) \cite{sun2009}, wave kernel signatures (WKS) \cite{aubry2011} and learnable optimal spectral descriptors (OSD) \cite{litman2014}. Spectral CNNs, defined on graphs, were first introduced in \cite{Bruna2013}. The main drawback of this method is that it relies on prior knowledge of the graph structure to define a local neighborhood for weight sharing. Consequently, the idea of graph convolutions has been extended in \cite{Henaff2015,Kipf2016} by limiting the support size of the learned spectral filters, making them independent of graph topology. In \cite{masci2015,boscaini2016,monti2017} another approach is presented, which defines a new form of local intrinsic patches on point clouds and general graphs, where the weights parameterizing the construction of patches are learned. Graph attention networks \cite{velickovic2017graph} learn a functional mapping to define pairwise weights based on the concatenated features of the involved nodes. The localized spectral CNN (LSCNN) \cite{boscaini2015}, which derives local patches from the windowed Fourier transform, can be seen as a combination of the spectral and the spatial method. \cite{bronstein2017} provides a comprehensive review of current research on this topic.

Deep learning applied directly on unordered point sets is considered in the PointNet framework \cite{qi2016,qi2017}. The input point set is recursively partitioned into smaller subsets and max pooling is used as a symmetric function to aggregate information regardless of point ordering.

Closest to our approach is the work of \cite{atwood2016}, that uses a power series of the transition matrix on a graph as diffusion operation to capture local node behavior, while we additionally employ multiple diffusion constants to build the filter kernels based on different variants of the normalized Laplacian. Moreover, we found it critically to build our network in a multi-layer fashion which was not considered in \cite{atwood2016}.

\subsection{Contribution}

In this work, we propose a simplified architecture that helps to overcome the limitations stated above, i.e. it can be employed for both grid and irregular graphs, has a comparable or better computational performance than classic CNNs and is theoretically connected to research on mean field inference approaches for graphical models in computer vision. As detailed in Section~\ref{secMethod}, we propose multi-kernel diffusion convolutional neural networks (mkdCNNs) based on two simple building blocks: isotropic, rotationally-invariant graph diffusion operators that propagate information across edges (on the graph) and trainable 1$\times$1 convolutions that manipulate features for each node individually. When employing multiple diffusion constants for the information propagation, which are linearly combined with the following 1$\times$1 convolution, powerful regional features, e.g. curvature, can be learned. A random walk approach is considered to further simplify the diffusion process. In Section~\ref{secExperiments} we successfully validate the proposed multi-kernel diffusion convolutional network on the tasks of learning pointwise correspondences between point clouds of different human poses as well as segmenting body parts.

\section{Multi-Kernel Diffusion CNNs for Point Clouds}
\label{secMethod}
\begin{figure}
\centering
\includegraphics[width=1.01\textwidth]{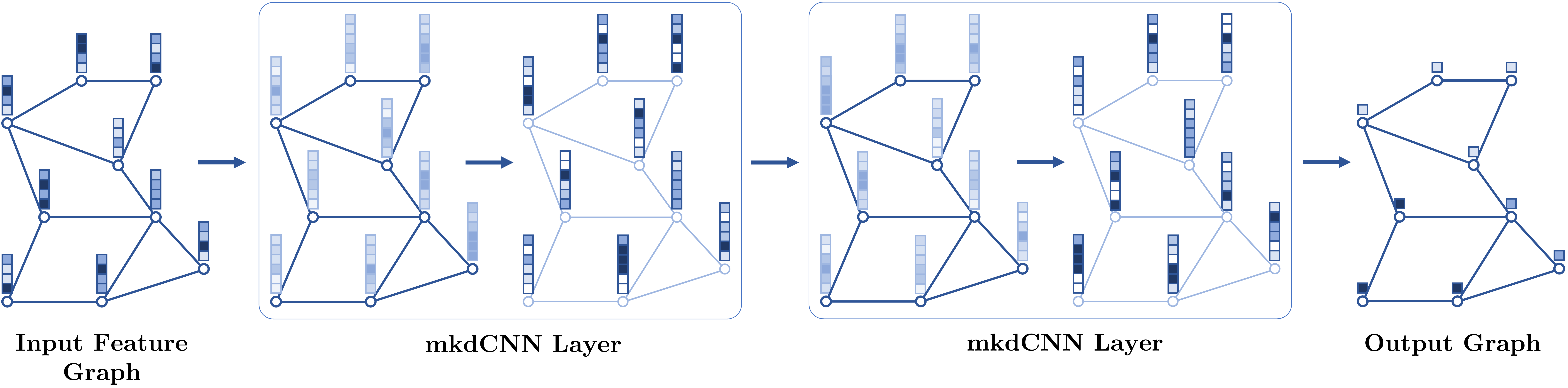}
\caption[]{Example of a two layer multi-kernel Diffusion CNN for node classification: Given an arbitrary input graph with $f$-dimensional features (left), we employ alternating layers of topology-independent diffusion operators with multiple isotropic kernels that propagate information across the graph, followed by $1\times1$ convolutions and activations that act on nodes individually and learn abstract representations of features (middle). In the end class predictions for each node a determined by a final $1\times1$ convolution (right).}
\label{figDCNN}
\end{figure}
\noindent
Input to our network is a matrix $\mathbf{P} \in \mathbb{R}^{n\times f}$, where the $i$-th row corresponds to one of $n$ points $\mathbf{p}_{i} \in \mathbb{R}^{f}$ of a point cloud in an $f$-dimensional feature space.

\subsection{Network Architecture}

Figure \ref{figDCNN} visualizes our proposed mkdCNN composed of the building blocks described below in detail. The layer input is a feature map defined on a graph. The weighted edges of the graph determine the feature propagation between nodes implemented as diffusion operation. The feature learning step consists of $1\times 1$ convolutions followed by non-linear activations. Therefore, its support is limited to each individual node. Stacked mkdCNN layers can be used in networks for global classification, with a final symmetric pooling function (e.g. max or average pooling), or for semantic node-wise segmentation in a fully convolutional manner.

\subsection{Input Feature Graph}

The simplest way to capture and represent local geometry in a point cloud is via a $k$-nearest neighbor graph $G_k$, where $\mathcal{N}_{k}(\mathbf{p}_i)$ denotes the set of the $k$-nearest neighbors of a point $\mathbf{p}_i$ and edge weights are defined by a distance metric $dist_{ij}$ between two points $\mathbf{p}_i$ and $\mathbf{p}_j$. An adjacency matrix $\mathbf{A}$ for the graph is constructed with entries
\[
    a_{ij} = 
\begin{cases}
    \frac{\exp(-dist^2_{ij})}{2 \cdot \sigma^2},& \text{if } \mathbf{p}_j\in \mathcal{N}_{k}(\mathbf{p}_i)\\
    0,              & \text{otherwise}
\end{cases}
,\]
where $\sigma$ denotes a scalar diffusion coefficient. In our work we employ multiple diffusion constants yielding different weighting schemes for the same graph. Spectral graph analysis \cite{Chung1995} allows us to extract further geometric properties from the point cloud, e.g. an intrinsic order of points, via the symmetric normalized graph Laplacian $\mathbf{L}_{\text{sym}}=\mathbf{I}-\mathbf{D}^{-1/2}\mathbf{A}\mathbf{D}^{-1/2}$. $\mathbf{I}$ denotes the identity matrix. The degree Matrix $\mathbf{D}$ is solely defined by its diagonal elements $d_{ii} = \sum_{j} a_{ij}$. For large point clouds it may be necessary to approximate the highly sparse matrix $\mathbf{L}_{\text{sym}}$ to maintain the computational efficiency of deep networks on GPUs. For this purpose, we can perform an eigendecomposition using only the first $m \ll n$ eigenvalues, such that
\[
	\mathbf{L}_{\text{sym}} = \mathbf{Q}\mathbf{\Lambda}\mathbf{Q}^{\intercal}
,\]
where the diagonal matrix $\mathbf{\Lambda}$ holds the $m$ eigenvalues and $\mathbf{Q}$ the corresponding eigenvectors. An alternative to the symmetric Laplacian is the random walk normalized Laplacian $\mathbf{L}_{\text{rw}}=\mathbf{I}-\mathbf{D}^{-1}\mathbf{A}$.

Input point features can be arbitrarily defined depending on the application and additional given information. For graphs derived from or based on regular grids like 2D images and 3D volumes such features may be simple grayscale values/patches or more suitable approaches, e.g. extraction of BRIEF descriptors~\cite{calonder2010brief}. Real world coordinates and surface normals can be extracted from 3D point clouds from stereo vision or time-of-flight systems. Once a graph is defined, the spectrum of the Laplacian itself can be used for feature extraction, e.g. B-spline based geometry vectors \cite{litman2014}. Furthermore, the construction of the mkdCNN makes it possible to learn meaningful information with no input features at all. In this case point features are simply initialized with ones.

\subsection{mkdCNN Layer}
Each of our proposed mkdCNN layers consists of two seperated steps: the diffusion operation and the feature learning.

\begin{figure}
\centering
\includegraphics[width=\textwidth]{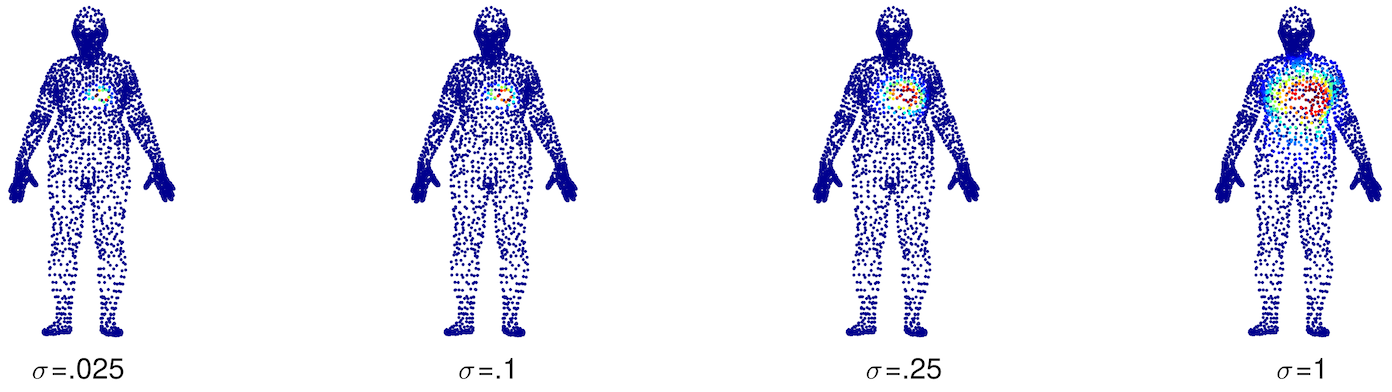}
\caption[]{Visualization of multiple isotropic diffusion kernels (for one point on the chest of a subject) employed in a single feature propagation step of our mkdCNN Layer.}
\label{figDiffusionKernels}
\end{figure}

To propagate features across the graph the Laplacian is used, thus making the propagation step for features independent of employed graph topologies and applicable to graph datasets with varying numbers of nodes. Essential to our mkdCNN layer is the use of multiple isotropic diffusion kernels as visualized in Figure~\ref{figDiffusionKernels}. Together with the following node-wise feature learning, expressive regional features with different local support can be extracted from the non-linear combination of all kernels. The diffused point cloud values $\mathbf{P}'$ can be computed as the solution of the diffusion process
\[
	\mathbf{P}' = (\lambda\mathbf{L}_{\text{sym}}+\mathbf{I})^{-1}\mathbf{P}
,\]
where $\lambda$ denotes the diffusion time \cite{desbrun1999implicit}. Approximating $\mathbf{L}_{\text{sym}}$ with few eigenvectors as mentioned above yields an efficient computation, as
\[
	\mathbf{P}' = \mathbf{Q}(\lambda\mathbf{\Lambda}+\mathbf{I})^{-1}\mathbf{Q}^{\intercal}\mathbf{P}
.\]
Therefore, diffusion is mainly affected by the parameters $k$, $\sigma$ and $\lambda$, that give control over the locality of the feature propagation. As our network can be trained in an end-to-end manner those parameters can either be learned or determined on a holdout validation set. As an alternative diffusion operation, that does not involve the costly matrix inversion, we also considered a random walker, such that
\[
	\mathbf{P}' = (\mathbf{I} - \mathbf{L}_{\text{rw}})^{t}\mathbf{P}
.\]
In this case the diffusion parameters are $k$, $\sigma$ and the number of diffusion steps $t$. Parallels to conditional random fields (CRFs) can be drawn. Our diffusion operation corresponds to one message passing step with the difference that the approximate mean and variance of features are propagated instead of an exact inference of all variables as in CRFs. In \cite{krahenbuhl2011efficient} a similar approach for efficient and approximate inference on grid-graphs is proposed that involves convolving a downsampled set of message variables with truncated Gaussian kernels.

In our proposed network, features are solely learned through $1\times1$ convolutions followed by a non-linearity. Besides adding depth to the network this choice is based on the analogy of our design with CRFs, where a label compatibility function is learned to penalize the assignment of different labels to nodes with similar properties \cite{krahenbuhl2011efficient}. Note that in CRFs the dimensionality of signals residing on each node is limited to the number of output labels and thus the compatibility function is restricted to only learn interactions across few classes, whereas in our approach the compatibility is established between feature maps. Furthermore, the exclusive use of $1\times1$ convolutions would make it conceptually easy to incorporate well studied building blocks from recent deep learning literature such as dense or residual connections into our network. Instance normalization and dropout are used to stabilize training and we employ a block of two $1\times1$ kernels each.

\section{Experiments}
\label{secExperiments}

Our new method is evaluated in two experiments: point descriptor learning and semantic body parts segmentation. We make use of the publicly available FAUST dataset~\cite{bogo2014faust}, which consists of $100$ surface meshes of $10$ different subjects, each scanned in $10$ different poses. The 3D meshes have a resolution of 6890 vertices and point-wise correspondences between the shapes  have been semi-automatically established for all points. As we are only interested in the scanned point clouds, we do not consider the given triangulations in our experiments. Following \cite{boscaini2015} we split the dataset in a disjoint training (subjects $1$-$7$, $70$~shapes), validation (subject $8$, $10$~shapes) and test set (subjects $9$-$10$, $20$~shapes).

\subsection{Point Descriptor Learning}

\begin{figure}
\centering
\includegraphics[width=\textwidth]{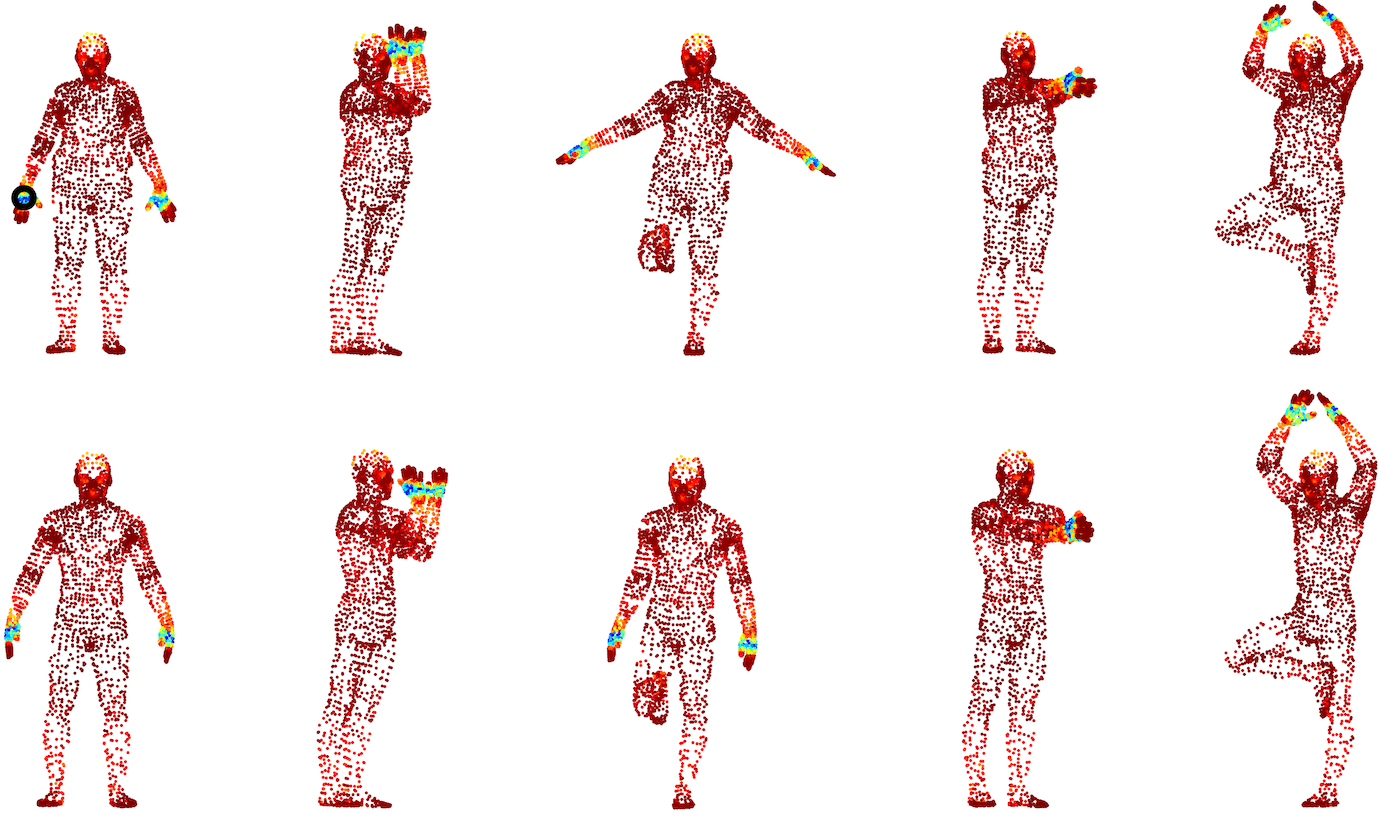}
\noindent\makebox[\textwidth]{\rule{\textwidth}{0.6pt}}

\vspace*{10pt}

\includegraphics[width=\textwidth]{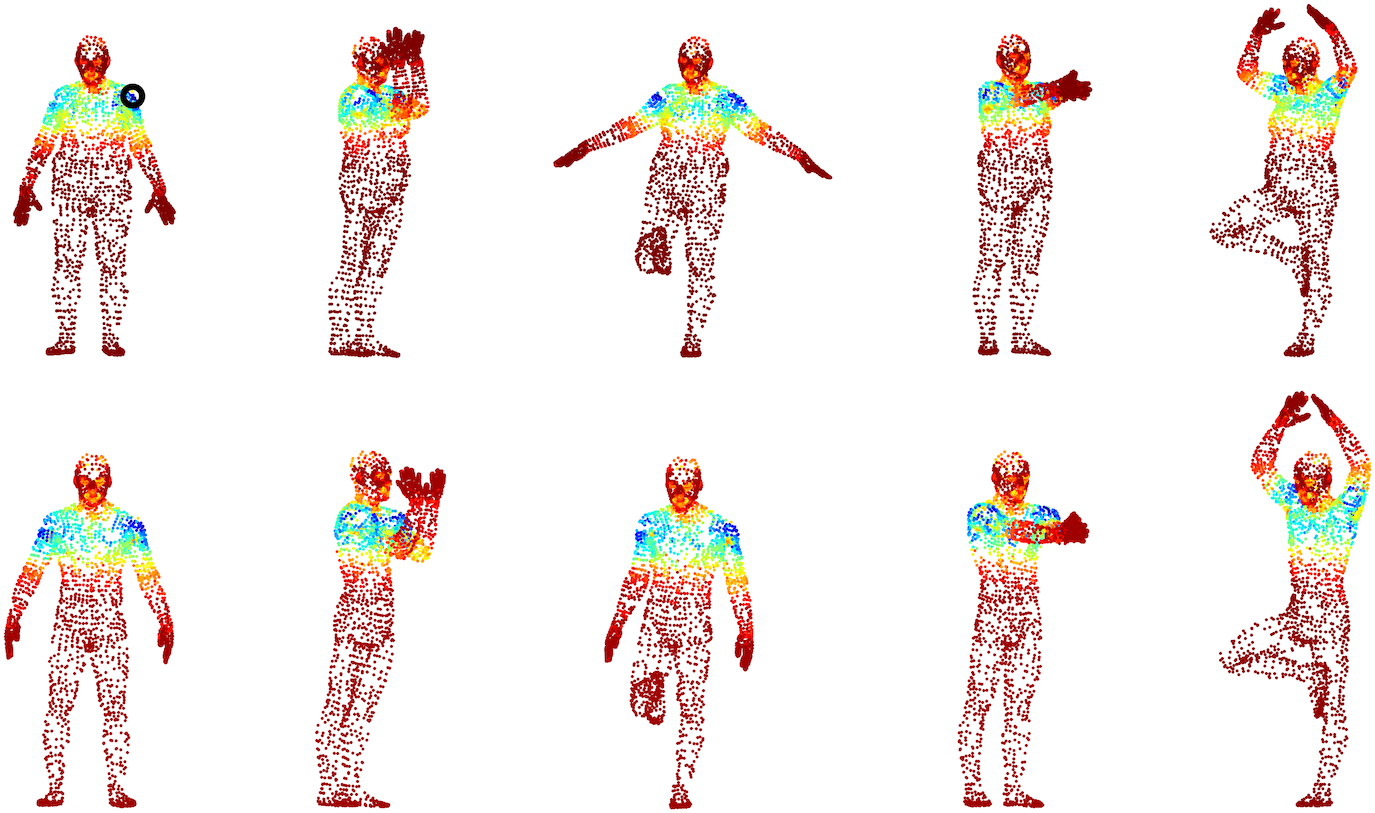}
\caption[]{Visualization of distances in the descriptor space between all points on a selection of shapes from the FAUST test set and a single point \circlemarker on a reference shape (upper left). Cold colors correspond to small distances. Distances are saturated at the median.}
\label{figDescriptorVisualization}
\end{figure}

\noindent
The graph Laplacian for all point clouds was computed using $k=100$ nearest neighbors. We employed a four layer mkdCNN using the random walk diffusion operation with parameters $\sigma=\{0.0125, 0.025, 0.05, 0.1, 0.125, 0.25, 0.5, 1\}$, $t=7$ and no features on the input graph. All parameters were chosen according to automatic hyperparameter optimization on the validation set. To train the descriptors we used a triplet hinge loss function - i.e. given a point on a randomly sampled shape its normalized Euclidean distance in the descriptor space to a non-corresponding point (on another random sampled shape) should be larger by a margin (here empirically set to $0.2$) than its distance to a corresponding point (on another randomly sampled shape). The descriptor dimension was set to $16$. Training was performed for $50$ epochs with the Adam optimizer \cite{kingma2014adam} and an initial learning rate of $10^{-4}$. For each optimization step we considered $6890$ triplets. We implemented our architecture in PyTorch \cite{paszke2017automatic} and train a model ($0.15$ million free parameters) on a Nvidia GTX 1070 8GB in around five hours. At test time the extraction of all $6890$ descriptors for one shapes takes approximately $5$ seconds. This time is dominated by the computation of the diffusion operation. Given precomputed diffusion operators our system is able to produce a throughput of $100$k points per second.

\begin{figure}
\centering
\includegraphics[width=.32\textwidth, height=.27\textwidth]{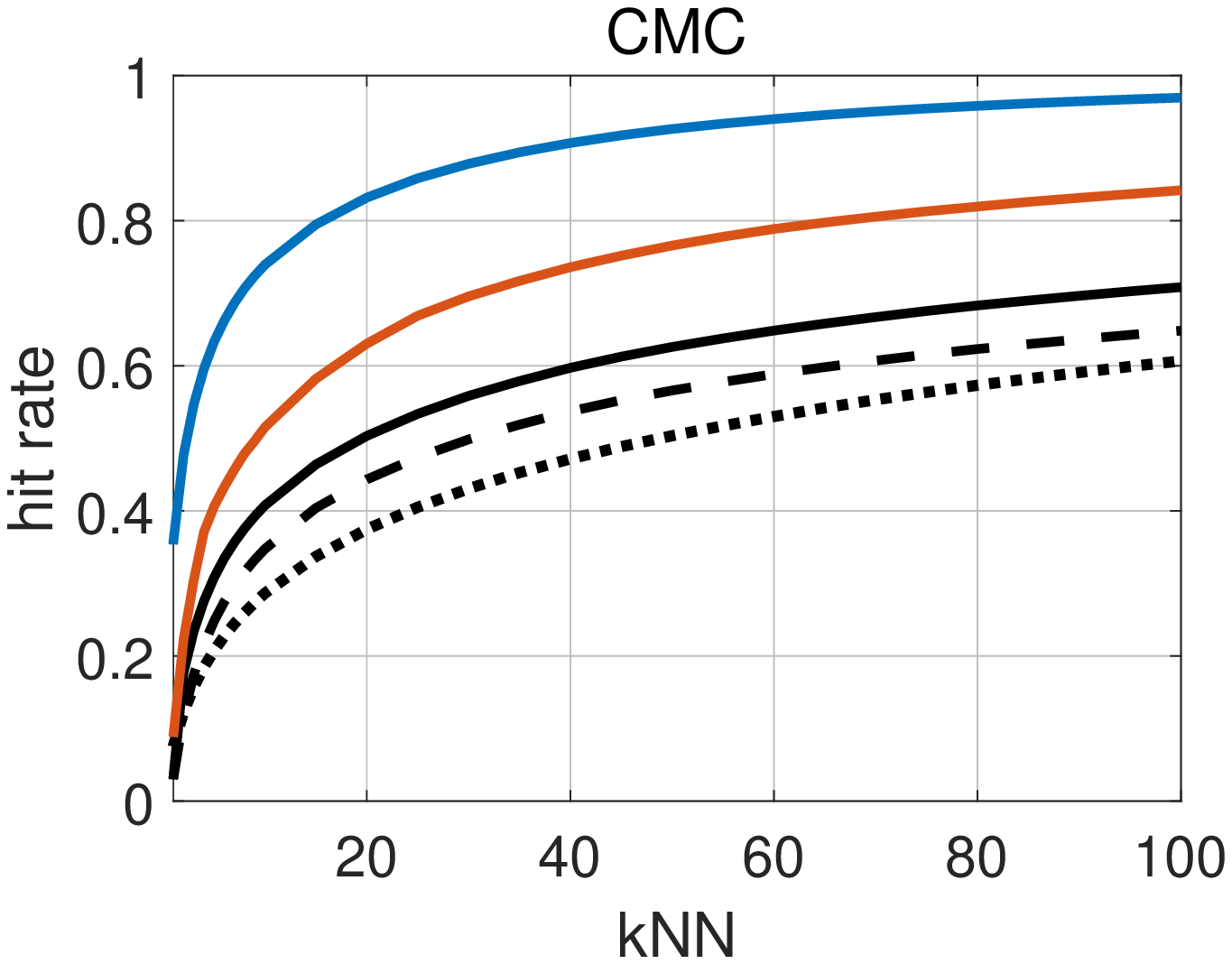}
\hspace{-2px}
\includegraphics[width=.32\textwidth, height=.27\textwidth]{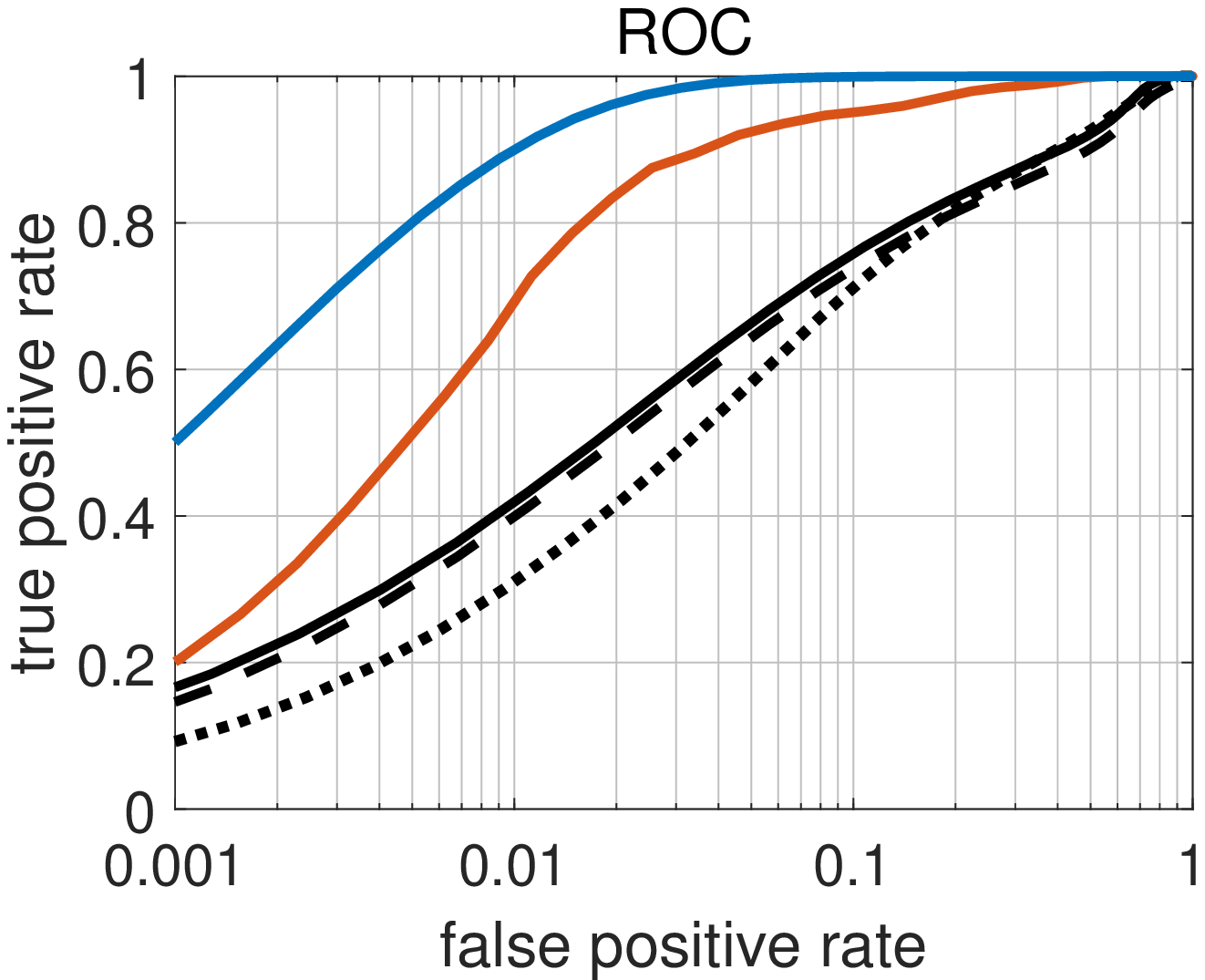}
\hspace{3px}
\includegraphics[width=.32\textwidth, height=.27\textwidth]{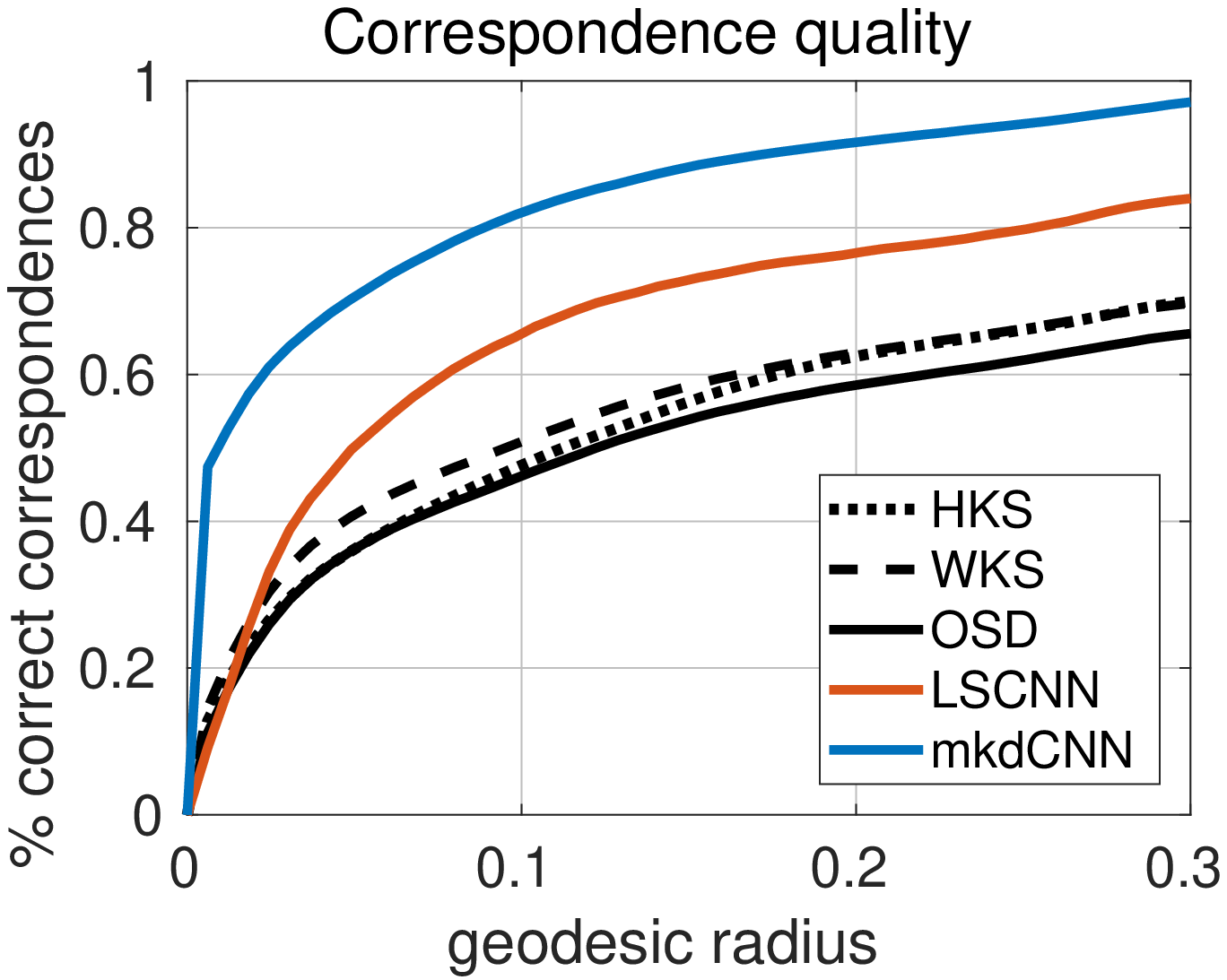}
\caption[]{Comparison of descriptor performances on the FAUST test set. For four comparison methods \boxmarker{colorBlack} \boxmarker{colorRed} and our approach \boxmarker{colorBlue} we report the cumulative match characteristic (CMC), receiver operating characteristic (ROC) and the correspondence quality, which measures the distance between matched and ground truth points on the underlying mesh of the point cloud.}
\label{figDescriptorResults}
\end{figure}

We compare the mkdCNN to four other spectral descriptor approaches, namely HKS \cite{sun2009}, WKS \cite{aubry2011}, OSD \cite{litman2014} and LSCNN \cite{boscaini2015}. Publicly available implementations of the approaches were used and parameters (e.g. $k$ for the computation of the graph Laplacian) optimized on the validation set.

\subsubsection{Results}

Figure \ref{figDescriptorResults} shows different evaluation results for all approaches on the FAUST test set. First, the cumulative match characteristic (CMC) is shown. It evaluates the retrieval performance by testing if the correct corresponding point on one shape can be found inside the next $k$-nearest neighbors from the set of all points of another shape. The $k$-nearest neighbors are determined by Euclidean distances in the descriptor space and the mean over all points and over all shapes is reported. The hit rate @kNN~$ = 10$ improved from $0.29$ (HKS), $0.35$ (WKS), $0.41$ (OSD) and $0.52$ (LSCNN) to $0.73$ (mkdCNN). The receiver operating characteristic (ROC) plots the true positive rate against the false positive rate of point pairs at several distance thresholds in the descriptor space. For a better distinction between the approaches, we plot the ROC curve in semilogarithmic scale. The measurements for the correspondence quality follow~\cite{kim2011blended}. The ground truth meshes are used to compute the geodesic distances between all points on a shape and the percentage of point pair matches that are at most $r$-geodesically apart from their corresponding ground truth points are reported. For the mkdCNN this means that over $80\%$ of point matches have a geodesic distance to their ground truth points of $10$~cm or less.

Figure \ref{figDescriptorVisualization} visualizes qualitative results of descriptors learned with the mkdCNN. A point is selected on a reference shape (on the right hand and left shoulder, respectively) and its distance in the descriptor space to all other points on the same and other shapes of the test set is computed. The distances are color-coded, where cold colors correspond to small distances. For most of the shapes distinct peaks around the ground truth are observable.

\subsection{Semantic Body Parts Segmentation}

The FAUST dataset does not include point-wise semantic labels for human body parts. Therefore, we labeled the points manually on a reference shape and transfered the labels to all other shapes via the known point correspondences. An exemplary ground truth labeling can be seen in Figure \ref{figDataDisturbances} (left). The $15$ labels correspond to the head, thorax, abdomen, left hand, left lower arm, left upper arm, left foot, left lower leg, left upper leg, right hand, right lower arm, right upper arm, right foot, right lower leg and right upper leg. The semantic segmentation on the FAUST dataset was also investigated in \cite{kleiman2018robust}, but only up to intrinsic symmetry (e.g. no distinction between right and left foot).

The default mkdCNN configuration for the semantic body parts segmentation is the same as for the descriptor learning task: the graph Laplacian is computed with $k=100$ nearest neighbors; we use the random walk \mbox{diffusion operation} with diffusion parameters $\sigma=\{0.0125, 0.025, 0.05, 0.1, 0.125, 0.25, 0.5, 1\}$ and $t=7$; no initial features are used on the input graph. For the classification task another $1 \times 1$ convolution is employed after the fourth mkdCNN layer producing softmax scores. The model is trained with a cross-entropy loss weighted with the root of inverse label frequencies and the Adam optimizer (initial learning rate: $10^{-4}$). Training is stopped after $50$ epochs.

\subsubsection{General Results}

\begin{figure}
\centering
\includegraphics[width=\textwidth]{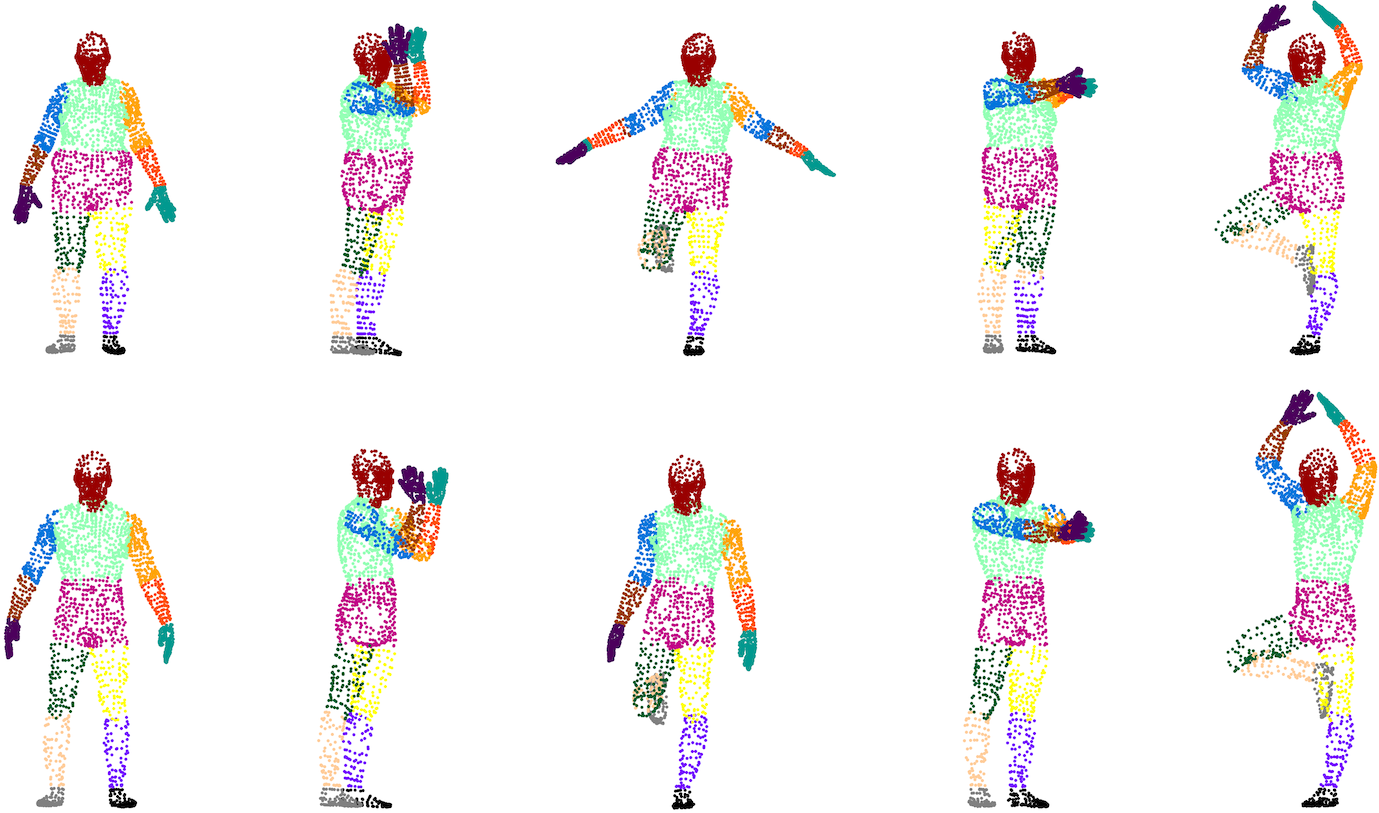}
\caption[]{Visualization of segmentation results on a selection of shapes from the FAUST test set. While the segmentation is visually convincing for most 3D point clouds, small inconsistencies can be observed. On the third shape in the top row points of the right lower arm \boxmarker{colorRLA}, right upper arm \boxmarker{colorRUA}, left lower arm \boxmarker{colorLLA} and left upper arm \boxmarker{colorLUA} are not always assigned to the correct side of the body. The same applies for points of the right upper leg \boxmarker{colorRUL} and left upper leg \boxmarker{colorLUL} on the fourth shape in the top row.}
\label{figSegmentationVisualization}
\end{figure}

Figure \ref{figSegmentationVisualization} depicts segmentation results for a selection of point clouds from the FAUST test set. The mkdCNN produces accurate and precise point cloud labels, even for challenging poses (fourth column: touching hands, fifth  column: right foot touches left knee). A Dice overlap of $0.95\pm 0.04$  (averaged over all labels and all shapes of the test set) confirms the good visual impression.

\subsubsection{Ablation study results}

\begin{figure}
\centering
\includegraphics[width=.32\textwidth, height=.27\textwidth]{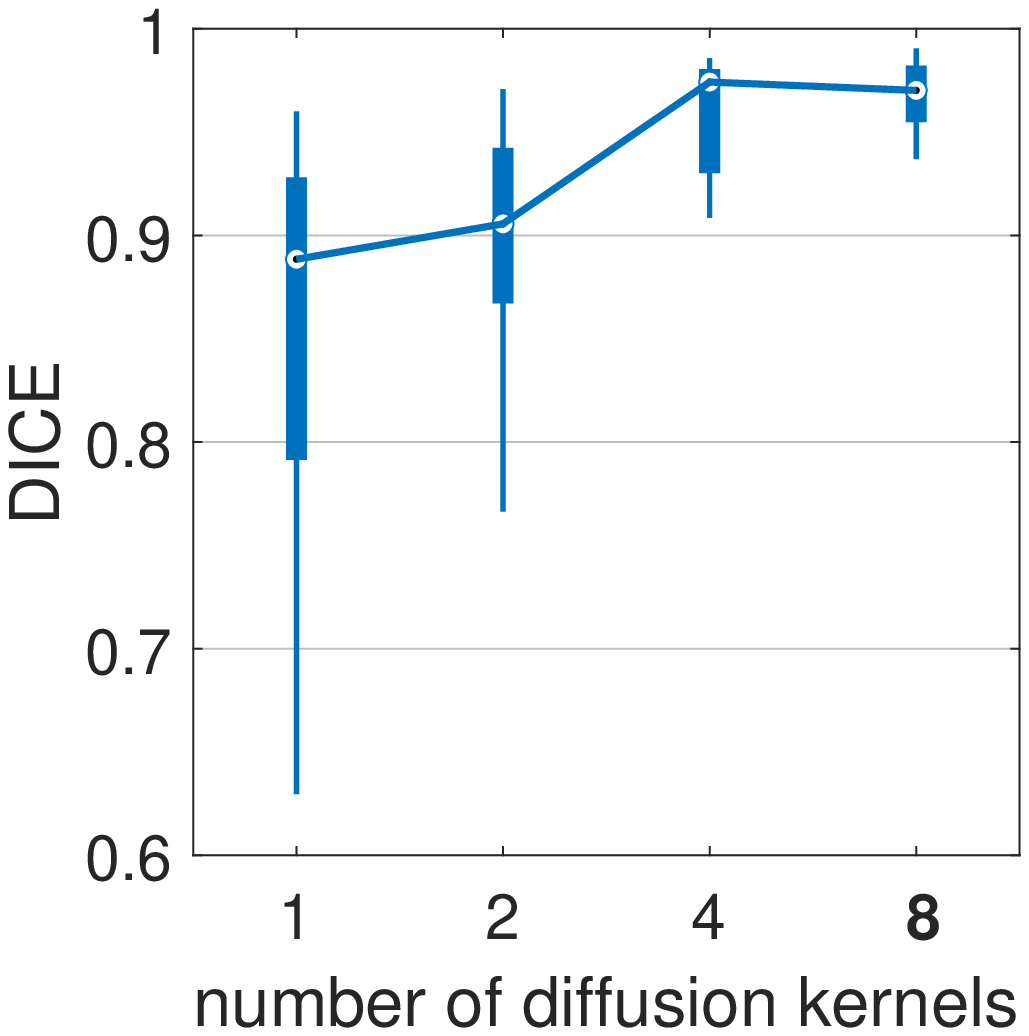}
\hfill
\includegraphics[width=.32\textwidth, height=.27\textwidth]{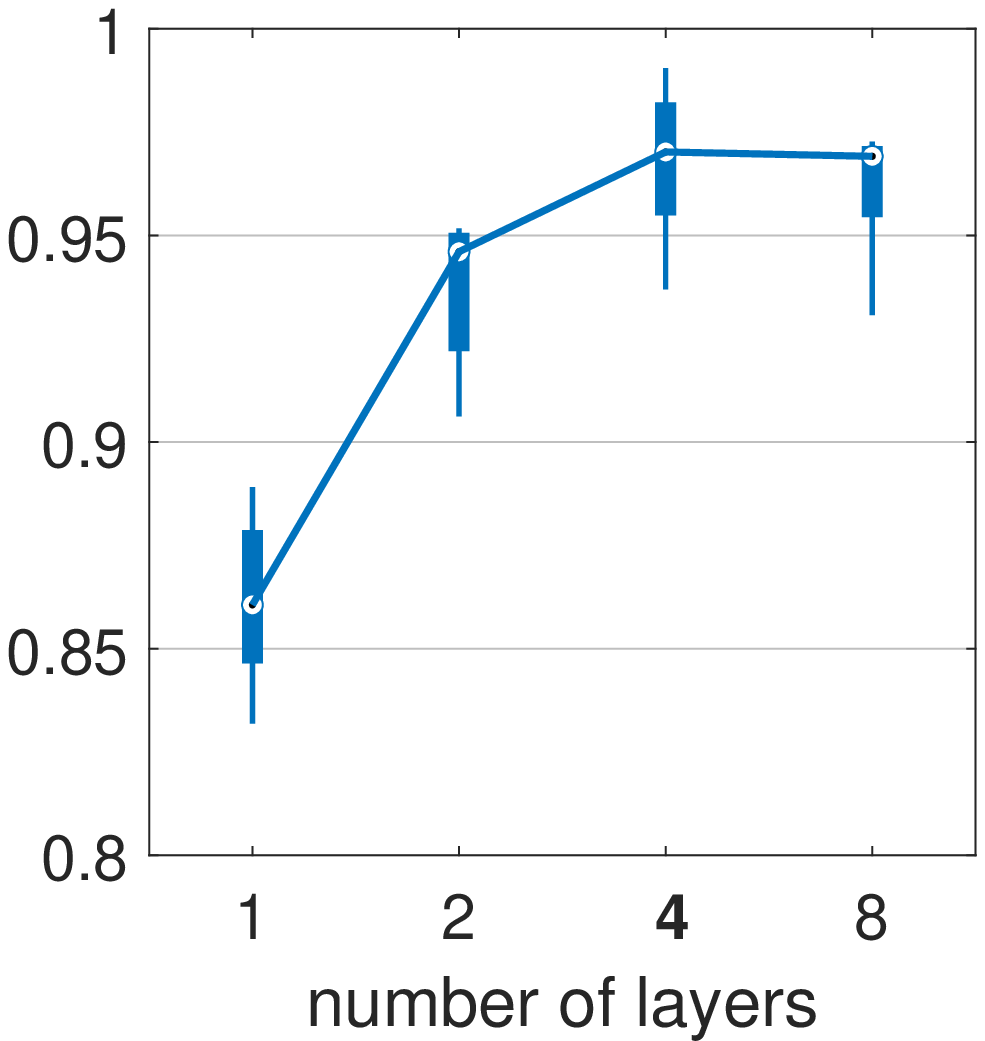}
\hfill
\includegraphics[width=.32\textwidth, height=.27\textwidth]{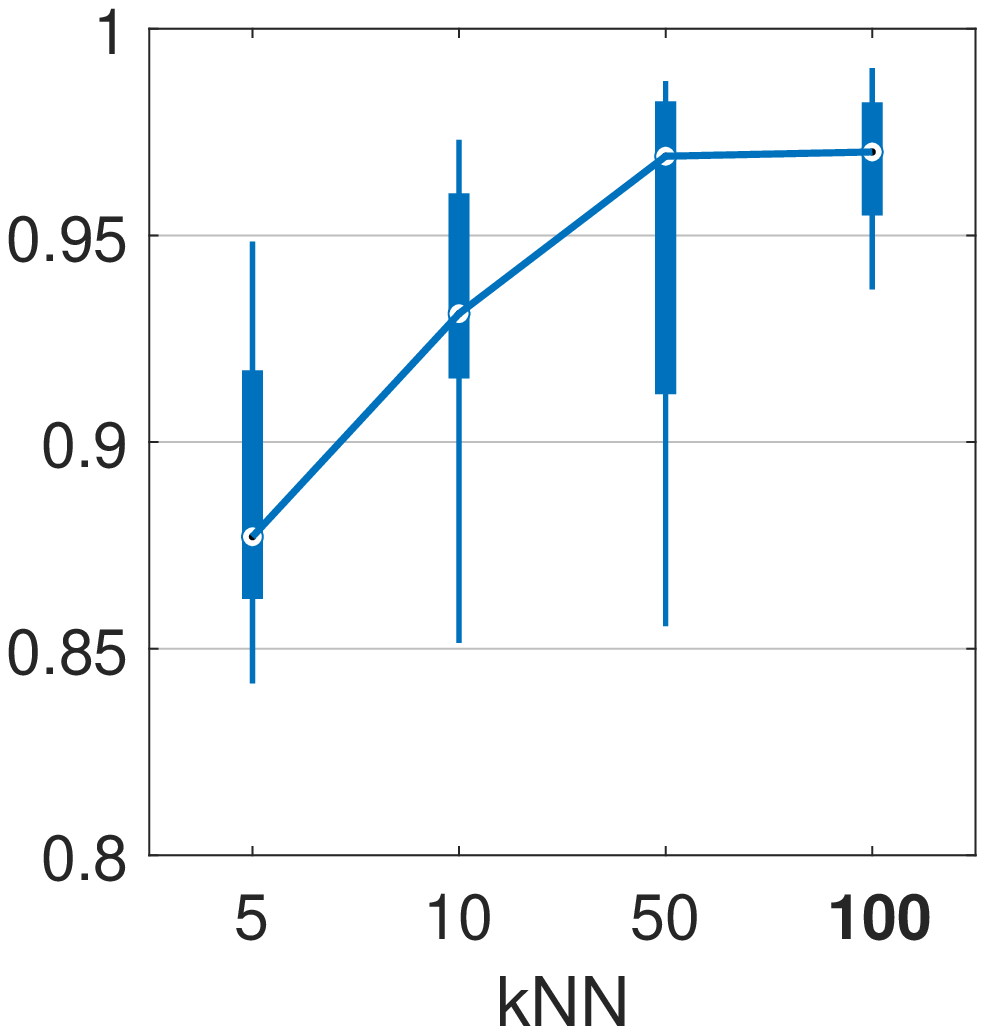}

\vspace*{15pt}

\includegraphics[width=.32\textwidth, height=.27\textwidth]{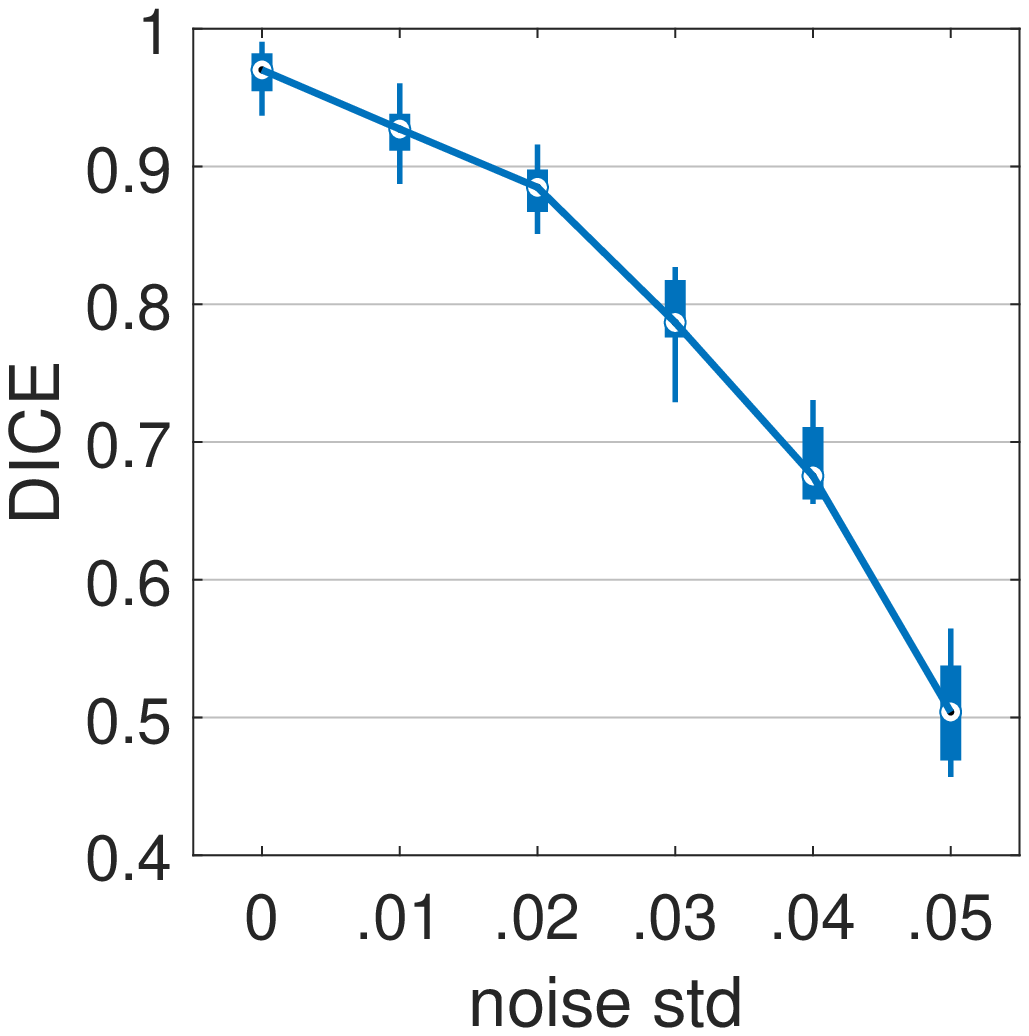}
\hfill
\includegraphics[width=.32\textwidth, height=.27\textwidth]{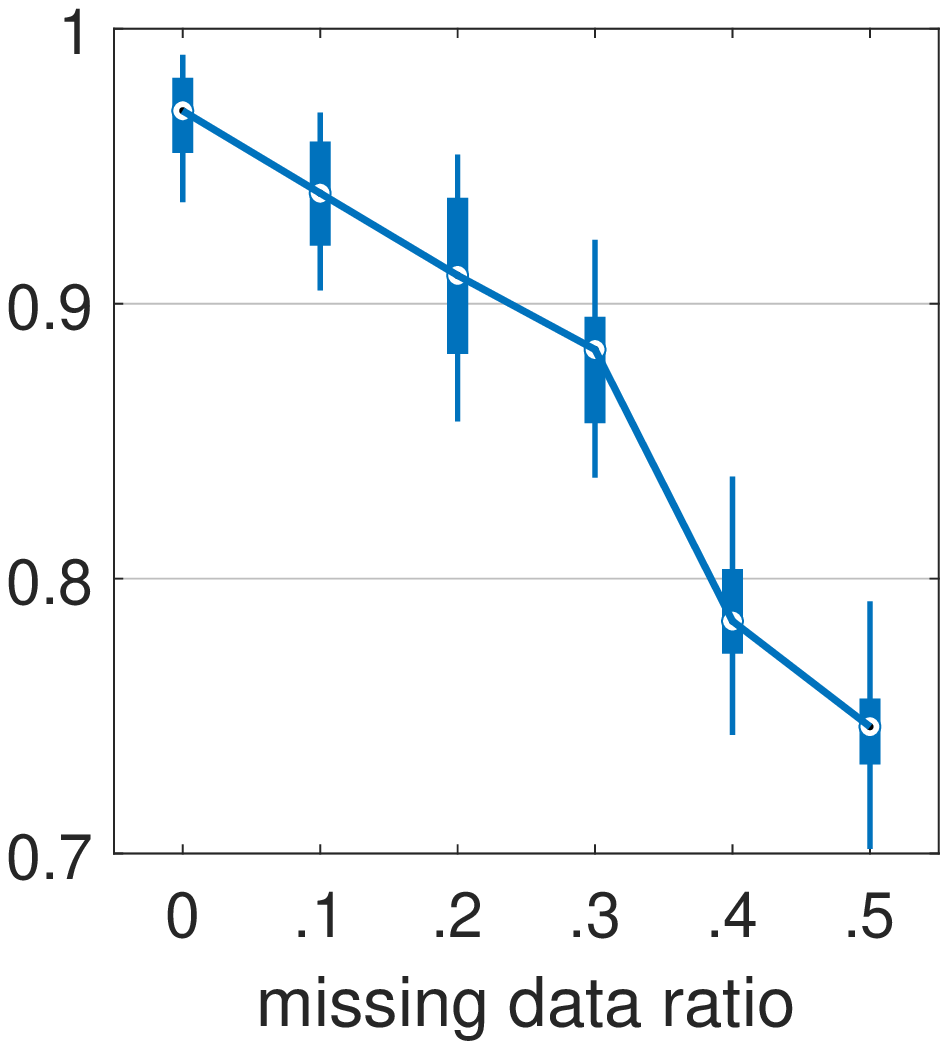}
\hfill
\includegraphics[width=.32\textwidth, height=.27\textwidth]{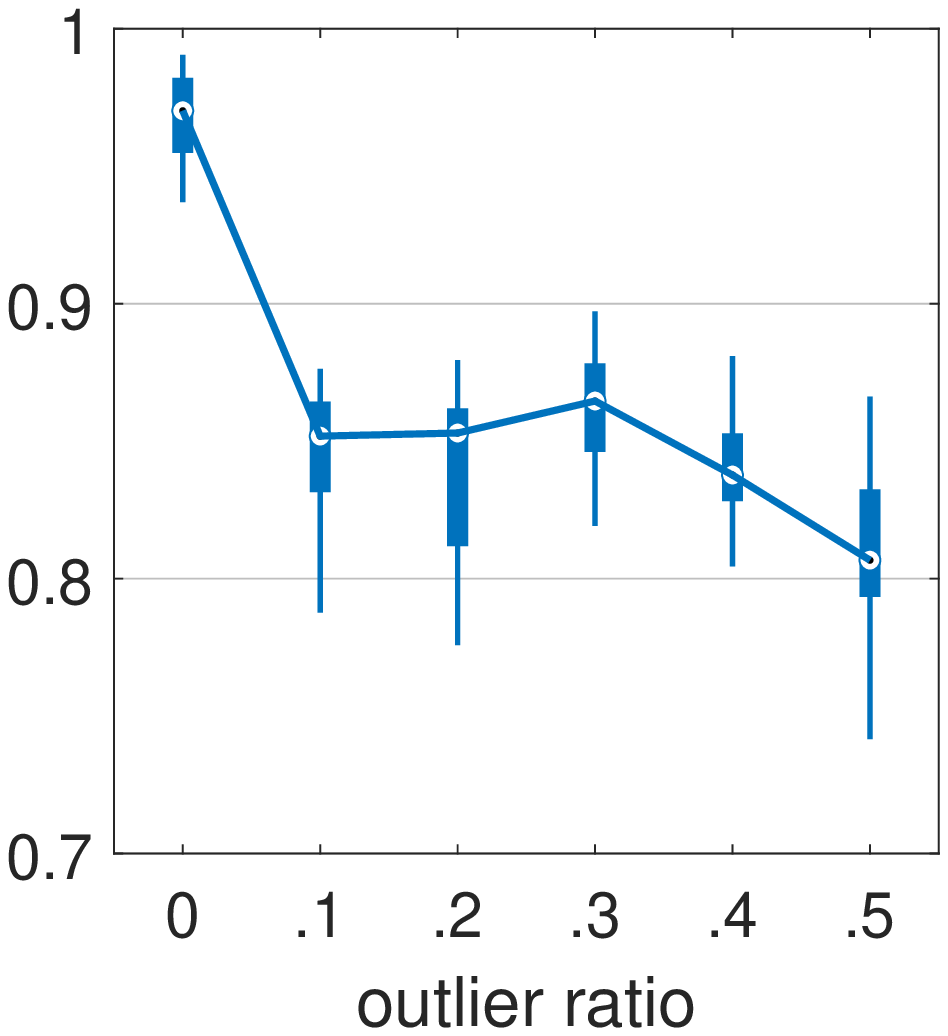}
\caption{Body parts segmentation results on the FAUST test set studying different parameter settings and data disturbances. Default configuration parameters are highlighted in bold.}
\label{figAblationResults}
\end{figure}

To understand the effect of different parameter and architectural choices in the mkdCNN, we perform several ablation experiments on the segmentation task. The default configuration for all experiments is the one described above. Using the exact diffusion process instead of the random walk approach yields a slightly improved Dice ($0.96\pm 0.03$ vs $0.95\pm 0.04$) at the cost of a much higher inference time (approximately $20$~s and $5.5$~s, respectively) due to the costly matrix inversion. Further evaluation results are shown as box-plots of Dice coefficients in Figure \ref{figAblationResults} (top row). In our first ablation experiment we study the effect of different number of diffusion kernels, i.e. the number of employed weighting schemes for the diffusion operation, on the segmentation results. Increasing the number of different $\sigma$ values from one to eight (and thus also increasing the total number of trainable weights of subsequent $1\times 1$ convolution layers from 40k to 150k) improves the mean Dice from $0.85\pm 0.11$ to $0.95\pm 0.04$. Particular interesting is the decreased standard deviation which implies a gain in robustness with respect to the variability between shapes. For the second experiment the number of mkdCNN Layers was set to $1$, $2$, $4$ and $8$, respectively. With an increased number of layers the size of the feature maps was reduced in order to keep the number of free parameters approximately the same for each configuration. Thus, the performance gain is introduced through a deeper mkdCNN architecture and not attributed to an increased capacity of the network. A difference in Dice overlap is especially recognizable between a one-layer and a two-layer mkdCNN. Another parameter that is not directly connected to the mkdCNN architecture but has a notable effect on the segmentation outcome is the number of nearest neighbors $k$ for the creation of the graph Laplacian from a given point cloud. For the mkdCNN it seems to be an advantage to be build on top of a graph with many locally highly interconnected nodes. The mean Dice coefficient increases from $0.88\pm0.04$ ($k=5$) to $0.95\pm 0.04$ ($k=100$).

\subsubsection{Robustness tests results}

\begin{figure}
\centering
\includegraphics[width=\textwidth]{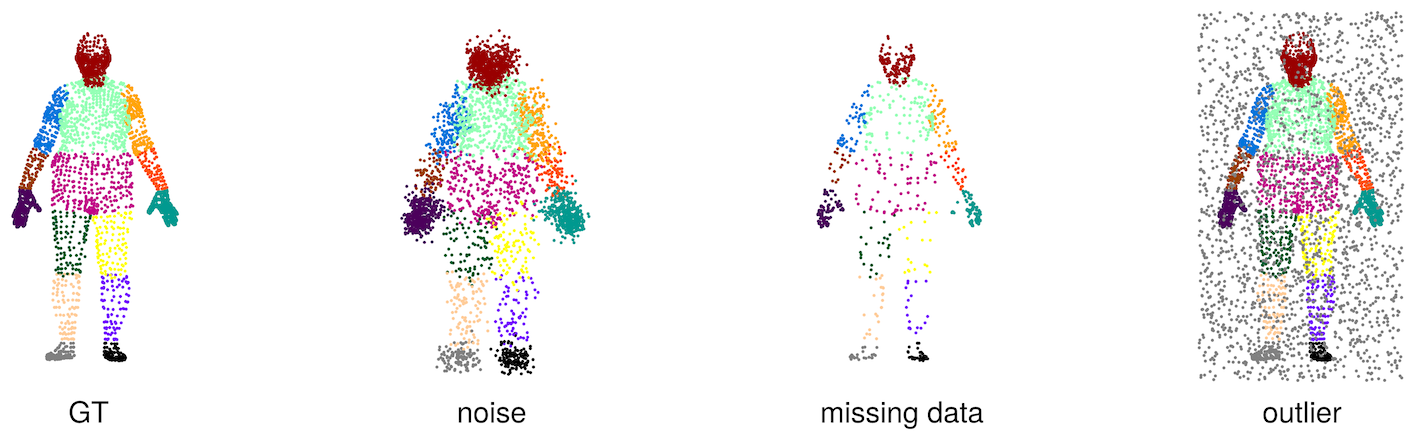}
\caption[]{Examples of our different data disturbance experiments. In this Figure Gaussian noise with a standard deviation of $0.03$ is added to the ground truth points. The ratio for missing data points as well as added outliers is $0.3$.}
\label{figDataDisturbances}
\end{figure}

A desirable property of a point cloud processing network is robustness against any disturbances of the input data. In a number of experiments we investigate the effect of different data disturbances on the segmentation results. Network parameters were not adapted for the robustness experiments. Figure \ref{figDataDisturbances} depicts the impact of the studied point perturbations on an exemplary ground truth point cloud. Robustness against noise is tested with random Gaussian noise added to the input cloud. We employ different standard deviations for the Gaussian (std~$=\{0.01, 0.02, 0.03, 0.04, 0.05\}$). With a std of $0.02$ the mean Dice coefficient is still approximately at $0.90$. The results deteriorates with a std of $0.03$ but Figure~\ref{figDataDisturbances} shows that the noise is already at a very high level and unresolvable ambiguities exists in this synthetic ground truth. In the next experiment we remove points at random with a certain ratio. Even if every second point is removed the mkdCNN produces segmentations with a mean Dice of $0.75$ without the need of adapting the network parameters. To investigate how the network can cope with outliers we randomly add points within the shapes bounding box. Added points are labeled as background. The results show that the segmentation task has become more difficult with the additional background class but is very robust against the ratio of outliers. Even for an outlier ratio of $0.5$, i.e. half of all points belong to the background class, the Dice overlap is above $0.80$. Figure \ref{figAblationResults} (bottom row) summarizes the results of all data disturbance experiments.

\section{Discussion}

Overall the results for both tasks are very promising and demonstrate a substantial improvement over both hand-crafted spectral features and graph convolution approaches. Especially the proposed use of multiple kernels and the consistent employment of mkdCNN layers in a multi-layer fashion helped to decrease error rates for Dice coefficients (i.e. $1-$Dice) in our investigated segmentation task by 66\% when using eight instead of a single diffusion kernel and by 52\% when increasing the depth from one to four layers. This is a significant improvement to the simple diffusion CNN in the work of \cite{atwood2016}, which is related to our \mbox{mkdCNN} in a configuration with only one kernel and a single layer. Despite using only topology-invariant and isotropic kernels, the learned non-linear combination in our proposed multi-kernel network help to create expressive and highly discriminative filters that enable accurate graph node classification.

When visually inspecting the point descriptor similarity in Figure \ref{figDescriptorVisualization} it appears that the learned 16-dimensional feature vectors do not differentiate well between symmetric structures (e.g. left and right shoulder). However, the semantic labeling tasks demonstrated that the subtle global differences in the human pose are sufficient to correctly label and distinguish between the right and left half of the body. For some rare cases the evident errors are indeed the inconsistent assignment of points to the correct side of the body (see Figure \ref{figSegmentationVisualization} top row, third and fourth column).

\section{Conclusion}

We have presented a new, simple architecture for descriptor learning and semantic segmentation on point clouds. By decoupling the graph propagation and feature learning step the mkdCNN overcomes the limitations of topology dependent approaches. Using the Laplacian and its approximation enables an efficient implementation of the diffusion of feature maps defined on sparse nodes that is transferable to different graphs and we showed that by providing multiple different kernels stronger features can be learned in each subsequent Layer. For the task of descriptor learning on point clouds from the FAUST dataset the mkdCNN (without any input features) shows better performance than a number of other spectral descriptors and learning approaches. Experiments on manually labeled body parts on the point clouds demonstrate the general feasibility of our approach for the task of semantic segmentation, even for highly noisy input (Gaussian noise, missing points, outliers). We validated several choices for our network architecture in ablation experiments and showed that a multi-layer mkdCNN with a high number of diffusion kernels build on top of a locally highly interconnected graph gives the best segmentation results in terms of Dice overlap. Visual inspection of the segmented point clouds expose rare failure cases due to ambiguities in the symmetry of the human body.

\section{Outlook}

Our mkdCNN framework provides some straightforward potential extensions for further improvements while maintaining its general design and inherent computational efficiency. Until now, we did not consider signals on our input point cloud but features like fast point feature histograms (FPFH) \cite{rusu2009fast}, RGB values (acquired with real-world 3D scanners like the Kinect) or spectral features can potentially increase the networks performance.

In this work we investigated the feasibility of the mkdCNN for learning on point clouds. As the diffusion operation is based on the graph Laplacian the network can be easily employed for general graphs. Testing our approach on graph datasets like Cora or PubMed \cite{sen2008collective} may yield interesting new insights.

An interesting research direction in general is to enable the possibility to not only learn features of a graph but also the connections (edge weights) between nodes and therefore incorporate mkdCNN into graph attention approaches \cite{velickovic2017graph}.

\bibliographystyle{splncs}
\bibliography{egbib}

\end{document}